\pdfoutput=1
\documentclass[conference]{IEEEtran}
\input{preamble}
 \StopCensoring

\usepackage{eso-pic}
\AddToShipoutPictureFG{%
  \AtPageLowerLeft{%
    \raisebox{.5\baselineskip}
      {\hspace{4.2\baselineskip}\footnotesize The final version of this paper is present in the proceedings of the $56^{\text{th}}$ edition of IEEE's International Symposium on Circuits and Systems (ISCAS), 2023.}%
  }%
}
\definecolor{darkred}{rgb}{0.7,0,0}

\title{Neural network scoring for efficient computing}

\author{
    \IEEEauthorblockN{\censor{Hugo Waltsburger\IEEEauthorrefmark{2}\IEEEauthorrefmark{3}, Erwan Libessart\IEEEauthorrefmark{3}, Chengfang Ren\IEEEauthorrefmark{2}, Anthony Kolar\IEEEauthorrefmark{3}, Régis Guinvarc'h\IEEEauthorrefmark{2}} }
    \IEEEauthorblockA{\protect\\ \censor{\IEEEauthorrefmark{2}SONDRA, CentraleSupélec, Université Paris Saclay, Gif-sur-Yvette}
    \\ \censor{\texttt{\{Hugo.Waltsburger, Chengfang.Ren, Regis.Guinvarch\}@centralesupelec.fr}}}
    \IEEEauthorblockA{\censor{\IEEEauthorrefmark{3}Université Paris-Saclay, CentraleSupélec, CNRS, Laboratoire de Génie Electrique et Electronique de Paris,} \protect\\ \censor{91192, Gif-sur-Yvette, France.}\protect\\ \censor{Sorbonne Université, CNRS, Laboratoire de Génie Electrique et Electronique de Paris, 75252, Paris, France} \\ \censor{\texttt{ \{Hugo.Waltsburger, Erwan.Libessart, Anthony.Kolar\}@geeps.centralesupelec.fr}}}
}
\pgfplotsset{compat=1.17}
\begin{document}
\maketitle

\begin{abstract}
    
Much work has been dedicated to estimating and optimizing workloads in high-performance computing (HPC) and deep learning. However, researchers have typically relied on few metrics to assess the efficiency of those techniques. Most notably, the accuracy, the loss of the prediction, and the computational time with regard to GPUs or/and CPUs characteristics. It is rare to see figures for power consumption, partly due to the difficulty of obtaining accurate power readings. In this paper, we introduce a composite score that aims to characterize the trade-off between accuracy and power consumption measured during the inference of neural networks. For this purpose, we present a new open-source tool allowing researchers to consider more metrics: granular power consumption, but also RAM/CPU/GPU utilization, as well as storage, and network input/output (I/O). To our best knowledge, it is the first fit test for neural architectures on hardware architectures. This is made possible thanks to reproducible power efficiency measurements. We applied this procedure to state-of-the-art neural network architectures on miscellaneous hardware. One of the main applications and novelties is the measurement of algorithmic power efficiency. The objective is to allow researchers to grasp their algorithms' efficiencies better. This methodology was developed to explore trade-offs between energy usage and accuracy in neural networks. It is also useful when fitting hardware for a specific task or to compare two architectures more accurately, with architecture exploration in mind.

\end{abstract}

\section{Introduction} 
\label{introduction}

Comparing the performance of different software and hardware architecture in the field of deep neural networks is a challenging endeavor. To establish a benchmark for comparison between various architectures, one must identify relevant metrics comparable across a wide variety of architectures and systems. Those metrics must have an adequate level of precision. However, the literature shows that benchmarks seldom comprise metrics other than the time required for the execution, the accuracy of algorithms, and the number of parameters and necessary multiply-adds \cite{hernandez, googlenet, resnet, alexnet}. This approach is relevant in a paradigm where algorithms need to run faster and more accurately, with little regard to the marginal cost of increased performance. However, this approach is not optimal if the objective is to optimize the power consumption of the algorithm, be it out of ecological concern or due to constraints on the hardware available. Such purposes require specific metrics. We advocate that comparing systems should be done using more criteria. The aim is to balance out accuracy and efficiency, specifically in power efficiency. This goal can be achieved by using scores. One of the first scores in the literature, characterizing the trade-off between accuracy and complexity, was introduced in \cite{canziani_nn_analysis} as:

\begin{equation}
    Score = \frac{\text{Accuracy}}{\text{Number of parameters}}
    \label{eq:score_1}
\end{equation}

With the idea of representing the amount of accuracy captured by a single parameter. However, the number of parameters does not always correlate well with the complexity of a network \cite{enable_DL_mobile}. Especially, convolutional neural networks comprise few parameters but are computationally expensive. Another iteration in neural network scoring, NetScore, taking multiply-accumulates (MACs) into account, was thus introduced in \cite{netscore}. They propose Netscore, which they define as 

\begin{equation}
    Netscore = 20 \log_{10}\Bigg(\frac{\text{Accuracy}^2}{\sqrt{\text{MACs}\times \text{Parameters}}}\Bigg)
    \label{eq:netscore}
\end{equation}

Netscore offers a more comprehensive view of the accuracy-efficiency trade-off of a neural network and seems especially relevant for convolutional neural network scoring. To elaborate upon this work, we want to introduce something that better reflects efficiency - especially in terms of power efficiency and adequation between hardware and software.

This paper presents a new score that uses measurements rather than technical information on the network. We believe introducing measurements in neural network scoring is necessary to characterize a neural network's behavior accurately. To obtain the necessary metrics, we introduce Tub.ai \cite{snorkel}, a new open-source tool that provides researchers with more diverse and granular metrics on their systems. We believe this methodology can be used in many fields - autonomous vehicles/drones, spaceborne applications, high-performance computing - and not only in AI. Its use is measuring both software and hardware performance. This paper also presents a benchmark of our score computed on various NN architectures during inference. Scoring was realized on various hardware platforms.  \\

This paper will first provide an overview of the motivation behind our new score and the tools we use in section \ref{metric}. We will then take measurements during inference using Tub.ai in section \ref{examples}, before presenting the scores obtained by several state-of-the-art NN architectures in section \ref{scores}. 


\section{Proposed score and tooling}

\label{metric}

As seen in section \ref{introduction}, some methodologies for NN-scoring already exist. However, they are solely computed using the accuracy and parameters from the neural networks' technical sheets (MACs and parameters). When considering power efficiency, the literature shows \cite{enable_DL_mobile} that power consumption does not scale linearly with either multiply-accumulates or parameters, which are the more common estimators for complexity in deep learning. Moreover, while these scores provide an estimation of the trade-off between accuracy and complexity, they are not measures. 

Since our logic is not to maximize accuracy, the best neural network for a specific application will depend on the hardware. We aim to strike an optimum between accuracy and power consumption. Only the inference phase is considered here. What seemed the most important to us was (i) accuracy, (ii) speed of inference, and (iii) power consumption per inference (in mWh). After several tests, we chose the formula in equation \eqref{eq:composite_score}: 

\begin{equation}
Composite\ Score  = \frac{\text{Accuracy}^2}{\text{Power consumption per inference}}
    \label{eq:composite_score}
\end{equation}
This formulation, when developed as 
\begin{equation}
\notag
Score = \frac{\Big(\frac{\text{Number of correct inferences}}{\text{Number of inferences}}\Big)^2}{\Big(\frac{\text{Total Power consumption}}{\text{Number of inferences}}\Big)}
\label{eq:composite_score_developed}
\end{equation}
Simplifies as 
\begin{equation}
   Score =  \frac{\text{Accuracy}}{\footnotesize{\text{{Average power consumed to obtain one correct inference}}}}
   \label{eq:physical_sense_score}
\end{equation}

We found that the power consumption per inference was sufficiently dependent on the speed of inference (the slower the inference, the higher the power consumption per inference) not to include the speed of inference as is. Using the rate of inference seemed to favor fast neural networks too much. This version of the score is easily comparable and has meaning from a physics standpoint. As shown in equation \eqref{eq:physical_sense_score}, it represents the accuracy captured by the average amount of energy consumed to obtain a single correct inference. 

As stated above, this score requires measurements made during inference. Therefore, a way of obtaining reliable measurements of energy consumption was needed.

One of the main problems encountered when attempting to measure power consumption is how to measure metrics on our systems precisely. To do this, we created a new tool, Tub.ai. Using various software and protocols, Tub.ai allows us to gather utilization metrics from our machines. Here, these metrics were gathered directly from the CPU\cite{rapl} and GPU\cite{snorkel} drivers, via RAPL and Nvidia-DCGM measurements.  

Once the data sources are set up, we need to aggregate them to exploit said data efficiently. To do this, we use a time series database that will store our data and allow us to query it at will. Over this database, we add a display engine that runs in the browser for easy visualization.

\begin{figure}

\begin{tikzpicture}[font=\small,thick]
 
\node[draw,
    minimum width=2cm,
    minimum height=1cm] (block7) {System Metrics};
 
\node[draw,
    left=of block7,
    minimum width=2cm,
    minimum height=1cm] (block8) {CPU Consumption};

\node[draw,
    right=of block7,
    minimum width=2cm,
    minimum height=1cm] (block10) {GPU Metrics};

\node[draw,align=center,
    below=1.5cm of block7,
    minimum width=2.5cm,
    minimum height=1cm,] (block11) {Timeseries Database};
    
\node[draw,align=center,
    rectangle,
    below=1cm of block11,
    minimum width=2.5cm,
    minimum height=1cm,] (block13) {Display Engine};
    
\node[draw,align=center, dashed,
    rectangle,
    below left=0.6cm of block11,
    minimum width=2.5cm,
    minimum height=1cm,] (block14) {Custom Metrics};
 
\draw[-latex] (block7) -- (block11);
\draw[-latex] (block8) |- (block11);
\draw[-latex] (block10) |- (block11);

\draw[-latex, dashed] (block14) -- (block11);

\draw[-latex] (block11) -- (block13);
 
\end{tikzpicture}
    \centering
    \caption{Flowchart of Tub.ai's architecture}
    \label{snorkel_flowchart}
\end{figure}
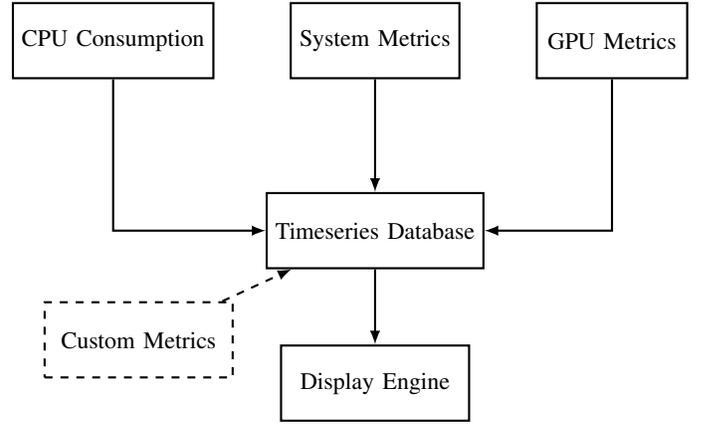

Once all the individual bricks have been aggregated, we obtain the architecture displayed in figure \ref{snorkel_flowchart}. Tub.ai only uses open-source applications and will be made open-source for the community to use after publication. Tub.ai is highly modular and can be used to gather an extensive range of data pertaining to the usage of a computer, including custom metrics. 

\section{Curating data on different NN architectures}
\label{examples}

\begin{table*}[!hbt]
\caption{Comparison of NN architectures based on Tub.ai's metrics, inference on ILSVRC2012 validation dataset, Nvidia A100} \label{tab:efficiency}
\setlength\tabcolsep{0pt} 
\footnotesize\centering

\smallskip 
\begin{tabular*}{\textwidth}{@{\extracolsep{\fill}}ccccccccc}
\toprule
\toprule
  Model Name  & Year & Validation  & Inference & Average GPU power & Total power & Average & Average GPU & Average \\
   &  & accuracy & time (s) & consumption (W) & used (Wh) & GPU usage (\%) & memory usage (\%) & CPU usage (\%)\\
\midrule
\midrule
  Inception ResNet V2 \cite{inceptionresnetv2}& 2016   & 80.3\% & 76  &211 & 4.5 & 92 & 99 & 4.3\\
  \midrule
  NasNet Mobile \cite{nasnet}& 2017   & 74.4\%     & 39     & 144 & 1.6 & 57 & 99 & 5.3\\
  \midrule
  NasNet Large \cite{nasnet}& 2017 & 82.5\% & 204 & 264 & 15 & 95 & 99 &2.7 \\
  \midrule
  MobileNetV3 Small \cite{mobilenetv3}& 2019 & 68.1\% & 15 & 107 & 0.5 & 40 & 99&10.7\\
  \midrule
  MobileNetV3 Large \cite{mobilenetv3}& 2019 & 75.6\% & 16 & 162 & 0.7 &62 &  99&9.9\\
  \midrule
  EfficientNetV2 S\cite{efficientnetv2}& 2021   & 83.9\%     & 89     & 253 & 6.3 & 88 & 99&4.7 \\
  \midrule
  EfficientNetV2 M\cite{efficientnetv2}& 2021   & 85.1\%     & 212     & 259 & 15.3 & 93 & 99 &3.4\\
  \midrule
  EfficientNetV2 L\cite{efficientnetv2}& 2021   & 85.7\%     & 365     & 262 & 26.5 & 96 & 99&3.2\\
  \midrule
\bottomrule
\end{tabular*}
\end{table*}

In this section, we extract data from our system while several computer vision NN architectures infer on the ILSVRC2012 dataset \cite{ILSVRC15}. This dataset was chosen due to its size and the amount of research available on it. It comprises 1000 classes, with a training dataset containing 1.28 million images, a test dataset containing 100,000 images, and a validation dataset containing 50,000 images. We chose to test several versions of the Inception-ResnetV2 \cite{inceptionresnetv2}, NasNet \cite{nasnet}, MobileNetV3 \cite{mobilenetv3} and EfficientNetv2 \cite{efficientnetv2} architectures. These architectures, released between 2016 and 2021, have either achieved then state of the art performance or approached it closely.\\

The experiment consists of inferences on ILSVRC2012. All models were implemented using Tensorflow 2.10 with Keras, Cuda 11.4, CudNN 8.3, and Python 3.10. We used several different testing machines representing several grades of computers.

\begin{itemize}
    \item For HPC servers: A single A100 GPU with 40GB of V-RAM, coupled with an AMD EPYC 7742 Processor (128 cores) and 504GB of RAM
    \item For upper-grade workstations: A Bi-Xeon 5222 workstation with 64GB of RAM and a Quadro RTX 5000 GPU with 16GB of V-RAM.
    \item For lower grade machines: A laptop comprising an i7-8565 CPU and 8GB of RAM 
\end{itemize}

For the sake of readability, all runs will be given different tags: "A100" for the A100 runs, "Quadro" for the RTX 5000 runs, "Bi-Xeon" for the dual Xeon 5222 runs, and "i7" for the laptop runs.
Each neural network's implementation was downloaded with pre-trained weights via the Tensorflow API. 

\begin{figure}[!th]
    \centering
    \input{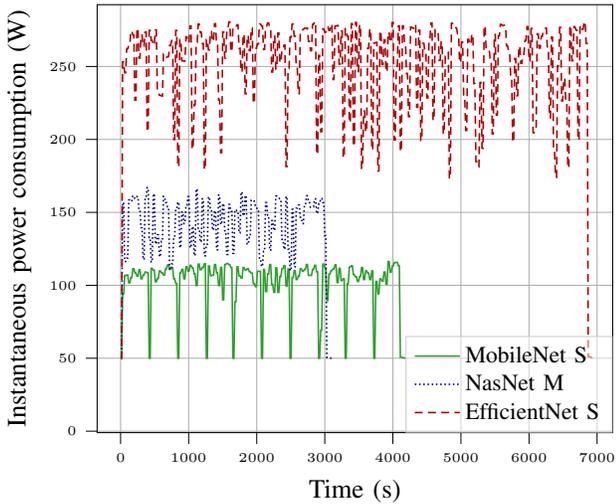}
    \caption{Nvidia A100 power consumption for 3 architectures}
    \label{fig:consumption_chart}
\end{figure}

Figure \ref{fig:consumption_chart} presents the instantaneous power consumption measured by Tub.ai for the A100 GPU during several inferences runs on the training dataset using three different architectures. EfficientNet S and NasNet Mobile were run three times in succession with no timeout. MobileNet Small was run ten times with a 30 seconds timeout between each run to provide a reference. It can be seen that the instantaneous power consumption may exhibit significant variations over a single run. It is interesting to observe that NasNet Mobile and MobileNet Small yield a much lower power consumption than EfficientNet. This suggests that the former architectures are bandwidth-bound on this setup, while the computing capabilities of the GPU bind the latter architecture. \\
In table \ref{tab:efficiency}, we provide the data obtained on the Nvidia A100 tests. All key indicators were measured during three inferences on the complete ImageNet validation dataset. Using three inferences was deemed robust enough as the results proved very consistent: even when chaining ten successive inferences over 1 million images each, the average deviation to the mean for each run settled below 2\%. The most significant observed outlier remained within 5\% of the average across all runs. No occurrences of thermal throttling or startup lag were observed. 

CPU usage can be considered a good indicator of the model's throughput for GPU runs since the CPU handles the I/O.
The CPU performance overhead introduced by Tub.ai is negligible compared to the CPU resources used by the inference during the benchmark: we measured an average load of 0.05\%. The GPU's constant 99\% memory usage is due to TensorFlow's inner workings, systematically reserving as much V-RAM as possible. When comparing the least and most consuming architectures, we can remark that there is a more than two-fold factor between instantaneously available metrics on this setup. We also observe a fifty-fold total power consumption factor. 




\section{Scoring neural networks}

\label{scores}





This metric aims to be a tool to evaluate the efficiency of different neural network architectures. It can be helpful in constrained systems, where there are limitations on both power consumption and computing power. It can also be used to evaluate the main leverage of any advancement in neural network architectures. The edge of an architecture may be its capacity to leverage Moore's Law and the algorithmic advances made on the various frameworks \cite{hernandez}, or it can be a less quantifiable change made in the structure of the architecture that makes it sparser, faster or lighter. 

Finally, other researchers may want to employ different coefficients or consider more factors for specific use cases. We hope that our approach and the developed tooling will be valuable for the community to introduce new scores. Using our score, we obtain the ranking presented in table \ref{tab:score_ranking}-and table \ref{tab:power_efficiency}. \\
\vspace*{-\baselineskip}
{\renewcommand{\arraystretch}{1.2}
\begin{table}[H]
    \centering
    \caption{Ranking of NN architectures using different scores}
\label{tab:score_ranking}
\begin{tabular}{|c||c|c|c|c|}

\hline
  Model&Score\cite{canziani_nn_analysis}&Netscore&A100 score &Xeon score\\
\hline
\hline
  ResNetV2   & 5 &5 & 4 & 4\\
  \hline
  NasNet Mobile  & 3 & 3& 3&2\\
  \hline
  NasNet Large & 7& 7 &7&6\\
  \hline
 \rowcolor{lightblue}MobileNetV3 S &1&1  & 1&1\\
  \hline
  MobileNetV3 L &2 &2 & 2&3\\
  \hline
  EfficientNetV2 S & 4 & 4 &5&5\\
  \hline
  EfficientNetV2 M & 6 &6&6&7\\
  \hline
  EfficientNetV2 L & 8 & 8 &8& 8\\
  \hline
\end{tabular}
\end{table}
} 
\addtocounter{table}{-1}  
\begin{table*}[!hbt]
\caption{Power efficiency of neural network architectures based on our new metric } \label{tab:power_efficiency}
\setlength{\abovecaptionskip}{0cm}
\footnotesize\centering

\smallskip 
\begin{tabular*}{\textwidth}{@{\extracolsep{\fill}}cccccccccccc}
\toprule
\toprule
  Model  & Validation   & \multicolumn{4}{c}{Power consumption/inference (µWh)}  & Score \cite{canziani_nn_analysis}& NetScore & \multicolumn{4}{c}{Our Score} \\
          &accuracy                                             & A100 &Quadro& Bi-Xeon& i7 & && A100{\tiny /A100}&Quadro{\tiny /A100}&Bi-Xeon{\tiny /A100} & i7{\tiny /A100}\\
\midrule
\midrule
  ResNetV2 \cite{inceptionresnetv2}   & 80.3\%  &   89&290 & 3,017&3,961&1.44& 47&7.2{\tiny /1}&2.2{\tiny /3.3}&0.21{\tiny /34}&0.16{\tiny /45}\\
  \midrule
  NasNet Mobile \cite{nasnet}   & 74.4\% &   31 &63&302&580& 14&70.1&17.9{\tiny /1}&8.8{\tiny /2}&1.84{\tiny /8.9}&0.95{\tiny /19}\\
  \midrule
  NasNet Large \cite{nasnet} & 82.5\% &  299&792 &4,792&9,766&0.9&42.4& 2.3{\tiny /1}&0.86{\tiny /2.6}&0.14{\tiny /16}&0.07{\tiny /33}\\
  \midrule
  MobileNetV3 S \cite{mobilenetv3}&  68.1\% &  9 &16 &143&120& \textbf{28.4} & \textbf{83.1} & \textbf{51.5}{\tiny /1} & \textbf{29.0}{\tiny /1.8} & \textbf{3.24}{\tiny /16} & \textbf{3.87}{\tiny /13}\\
  \midrule
  MobileNetV3 L \cite{mobilenetv3}& 75.6\% &  14 &29& 334&293&14&74.5&40.8{\tiny /1}&19.5{\tiny /2.1}&1.70{\tiny /24}&1.95{\tiny /21}\\
  \midrule
  EfficientNetV2 S \cite{efficientnetv2}& 83.9\%  & 125&286 &3,340&4,195&3.9&54.2& 5.6{\tiny /1}&2.46{\tiny /2.3}&0.21{\tiny /27}&0.17{\tiny /34}\\
  \midrule
  EfficientNetV2 M \cite{efficientnetv2}&85.1\% & 305&877 &8,794&12,236&1.6&46.0 & 2.4{\tiny /1}&0.83{\tiny /2.9}&0.08{\tiny /29}&0.06{\tiny /40}\\
  \midrule
  EfficientNetV2 L \cite{efficientnetv2}& 85.7\%  & 530 &1,539& 16,698&24,346&0.7&39&1.4{\tiny /1}&0.48{\tiny /2.9}&0.04{\tiny /32}&0.03{\tiny /46}\\
  \midrule
\bottomrule
\vspace*{-8mm}
\end{tabular*}
\end{table*}

First, our ranking seems generally consistent with the scores obtained using \cite{canziani_nn_analysis} and \cite{netscore}. Second, we observe a difference in the hierarchy depending on the setup: on the Bi-Xeon platform, NasNet Mobile outperforms MobileNetV3 Large, and NasNet Large outperforms EfficientNetV2 Medium. Third, by observing the factor of proportionality between our scores and the A100 score (in small fonts next to each of our scores), we can see that some architectures (Inception ResNetV2, EfficientNetv2 Large) are much more penalized than other architectures (MobileNet, NasNet) when changing setup. This result confirms the value of taking measurements on neural network inferences rather than relying on platform-independent metrics. We hypothesize that this difference is due to a balance between computing and bandwidth requirements making some architectures less hardware constrained. 

Despite having the highest average power consumption of all platforms, the A100 has the highest score across all architectures. While higher-grade hardware has higher idle and in-charge energy usage, it usually exhibits a lower power consumption per inference due to increased inference speed. Notwithstanding, while the bi-Xeon setup outperforms the i7 on most architectures, the i7 fares better on MobileNets. This result emphasizes the need to study the envisioned application of a given neural network to extract its maximum performance. It also calls for more work on the architecture of neural networks to identify the limiting factor across different hardware settings and obtain the best hardware-software fit.

MobileNetv3 Small has a net lead over all other networks and ranks first on all platforms, which seems reasonable since the MobileNet architecture was created to be \textit{"a class of efficient models [...] for mobile and embedded vision applications"} \cite{mobilenets}. Some of the more recent models perform worse than older models in terms of score. It seems likely that, in the future, the discrepancy between networks made for optimum efficiency and those created solely for performance will drift apart further. This observation leads to new approaches, with some recent architectures \cite{regnet} \cite{efficientnetx} seeking to optimize training/inference speed rather than accuracy or parameter efficiency and advocating for better architecture-algorithm adequation.

We hope that it will be more commonplace in the future that some benchmarks rank neural network architectures using not only the validation accuracy but also more metrics, especially power consumption measured on both training and inference.

\section{Conclusion and future works}

In this paper, we attempted to create a score based not on a technical datasheet but on measurements made on a platform. The idea is to measure the algorithm-architecture adequation between specific NNs and hardware. To obtain the required metrics to compute our score, we created Tub.ai. 

Tub.ai provides researchers with various metrics that can be leveraged to create meaningful comparisons between different software implementations and architectures. Tub.ai is open source, induces very little overhead in terms of performance, and is easy to use. This approach significantly benefits system developers who have to develop or optimize algorithms for specific hardware. 

Our deep neural network architecture benchmarks estimate how well they are fitted to be run on different platforms. The score values we have obtained across several architectures show that there can be significant discrepancies between platforms for the same neural network - which proves the interest of scoring using measurements. To our best knowledge, this is the first time power efficiency measurements have been included in the performance evaluation of neural network architectures. In a context of growing concern for the ecological impact of machine learning and high-performance computing in general, it is a step forward in the field of HPC where the main focus is not solely the algorithm's or architecture's performance but also their power efficiency.

In the future, we plan on furthering the work we have done on scoring by exploring other hardware, architectures, and frameworks. Using identical neural network architectures implemented on different frameworks and running them on various test hardware, we would like to evaluate how well available deep learning frameworks can exploit the hardware's capabilities. An examination of the power efficiency of specialized neural processing units with regard to our score is planned. Similar undertakings on other datasets are considered. We also plan on updating Tub.ai and making it more precise and easier to use. We would be especially interested in seeing the results of neural network architecture search using this new score as the optimizing criterion.
\printbibliography
\end{document}